%% file: IEEE-conference-template-062824.tex
\documentclass[conference]{IEEEtran}
\IEEEoverridecommandlockouts

\usepackage{amsmath,amssymb,amsfonts}
\usepackage{graphicx}
\usepackage{textcomp}
\usepackage{xcolor}
\def\BibTeX{{\rm B\kern-.05em{\sc i\kern-.025em b}\kern-.08em
    T\kern-.1667em\lower.7ex\hbox{E}\kern-.125emX}}

\usepackage{caption}
\usepackage[hidelinks]{hyperref}
\usepackage[justification=centering]{caption}
\usepackage{amsmath}
\usepackage{stfloats}
\usepackage{float}
\usepackage{booktabs}
\usepackage{siunitx}
\usepackage{amssymb}
\usepackage{multirow}
\usepackage{pgfplots}
\usepackage{cleveref}
\usepackage{subcaption}  
\pgfplotsset{compat=1.18}
\usepackage{algorithm}
\usepackage[noend]{algpseudocode}
\usepackage{tikz}
\usetikzlibrary{calc} 
\usetikzlibrary{shapes.geometric, arrows.meta, positioning,matrix}
\usetikzlibrary{positioning, arrows.meta}
\tikzset{
  node/.style={circle, draw=black, thick, minimum size=1.5cm, align=center},
  graynode/.style={circle, draw=gray, fill=gray!10, thick, minimum size=1.5cm, align=center},
  arrow/.style={-{Stealth}, thick},
}
\usepackage{makecell} 

\usepackage[style=ieee, maxbibnames=99, citestyle=numeric-comp, doi=false,url=false, natbib=true,natbib=true]{biblatex}
\AtEveryBibitem{\clearfield{issn}}
\AtEveryCitekey{\clearfield{issn}}
\begin{document}

\title{Importance Sampling and PCA for Finding\\ Failures in Commercial Autonomous Vehicles
\thanks{H. Warner*, D. Eddy, S. Parjan, C. Cahilly, 
H. Delecki, and M. J. Kochenderfer are with the Department 
of Aeronautics and Astronautics, Stanford University, 
Palo Alto, CA (*hlwarner@stanford.edu).}
\thanks{M. Kleinst\"{a}uber, C. Shinde, and J. Lopez  are with 
Torc Robotics, Blacksburg, VA.}}

\author{\IEEEauthorblockN{Hailey Warner, Duncan Eddy, Shreya Parjan,
Caroline Cahilly, Harrison Delecki, \\
Matthias Kleinst\"{a}uber, Chaitanya Shinde, Jerry Lopez,  Mykel J. Kochenderfer}}

\maketitle

\begin{abstract}
Methods for discovering rare failures in autonomous systems have so far been demonstrated almost exclusively in simulations with simple, academic driving stacks, leaving open whether they generalize to the more robust planners used in commercial systems. We address this gap by applying two rare-event discovery algorithms to a commercial autonomous trucking stack. Adaptive stress testing (AST) uses reinforcement learning to search for the most likely noise trajectories leading to a simulated collision, while diffusion-based failure sampling (DiFS) trains a denoising diffusion model to sample a diverse set of failures. We show that both algorithms find simulated collisions during merge and cut-in maneuvers where traditional Monte Carlo simulation does not. To make these failures actionable, we introduce a statistical analysis based on principal component analysis (PCA) that classifies failures into common modes and identifies the timesteps that most influence the outcome. We cluster the principal components and invert the PCA transform to recover generalized noise trajectories, and show that these trajectories reproduce failures in identical and similar scenarios. This provides a path from failure discovery to systematic diagnosis of perception-level flaws.
\end{abstract}


\input{Sections/intro}

\input{Sections/methodology}
\input{Sections/results}
\input{Sections/conclusion}
\input{Sections/acknowledgement}

\renewcommand*{\bibfont}{\footnotesize}
\printbibliography

\end{document}

%% file: Sections/intro.tex
\section{Introduction}


Failures in autonomous vehicles are rare but potentially catastrophic. Real-world testing is too dangerous and expensive to rely on for comprehensive coverage, and while Monte Carlo simulations offer an alternative, they are time consuming and may never surface the most severe failure cases \cite{koopman}. Learning-based importance sampling algorithms have been developed to address this challenge, however they have so far only been demonstrated on simpler, academic autonomous driving planners such as the Intelligent Driving Model (IDM)\cite{ast_av}. The question lingers whether they remain effective at finding failures in more robust, commercial planners. Here, we apply Adaptive Stress Testing (AST) and Diffusion-based Failure Sampling (DiFS) for the first time to a commercial autonomous trucking system to demonstrate that they successfully find rare failures in a sample-efficient manner for real-world autonomous driving systems where Monte Carlo testing does not find failures. The results show that AST and DiFS are viable and practical tools for improving the safety of deployed autonomous vehicles.

Importance sampling techniques excel at finding rare, unexpected failures in autonomous systems \cite{neurips}. AST has been shown to discover such failures by manipulating perception noise to induce collisions with vehicles or pedestrians \cite{moss}. DiFS has shown similar promise on toy problems \cite{difs}, but both algorithms have only been validated on simplified open-source driver models. Commercial planners are far more robust, so failures are rarer and harder to sample. It remains an open question whether these algorithms can generalize to such planners.
\begin{figure}[t]
    \centering
\includegraphics[width=0.7\linewidth]{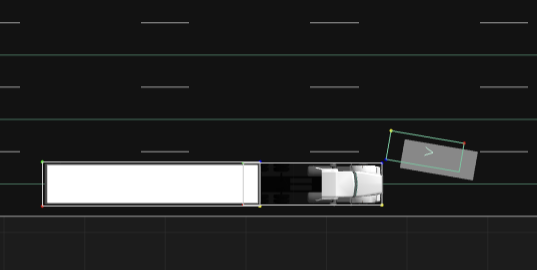}
    \caption{Cut-in scenario in Object Sim environment. Actor (green outline) cuts in front of ego (truck). The ego perceives the actor to be located at the noisy position (gray box).}
    \label{fig:cut-in}
\end{figure}
Additionally, raw noise inputs alone offer little actionable insight into why the system failed. Prior root-cause analyses have largely operated at the planning or environmental level: \citeauthor{rocas} performed sensitivity analyses over environmental and algorithmic variables to pinpoint external triggers and planning errors \cite{rocas}; co-occurrences of lighting, weather, and semantics with collision types have been cataloged in AV crash datasets \cite{kurse}; and PCA has been used on industrial processes to identify influential variables and learn Bayesian networks over them \cite{pca_bn}. None explore sensor noise profiles as direct diagnostic signals, leaving a gap between failure discovery and systematic failure diagnosis at the perception level.

To move beyond individual failure cases toward systematic diagnosis, we use principal component analysis (PCA) to characterize, summarize, and explain discovered failures found through importance sampling. PCA extracts a common low-dimensional noise structure across simulated AV failure cases. Principal components can be clustered into interpretable modes and traced back to specific flaws in the AV's perception system. We refer to such modes as ``eigenfailures." PCA additionally gives a measure of similarity across failure trajectories through the explained variance, and determines the timesteps with the highest influence on failure outcomes to focus remediation efforts. By clustering principal components with K-means, then inverting the PCA transform, we recover generalized noise trajectories that reveal underlying commonalities across many failures.

This paper offers two main contributions: (1) the first demonstration of both AST and DiFS on a commercial AV planner and (2) our PCA-based approach to systematically identify and reproduce canonical sensor failure modes in arbitrary driving scenarios. We demonstrate our approach on a highway cut-in scenario, in which an actor vehicle merges into the ego truck's lane while perception noise distorts the ego's estimate of the actor's position. Our results show that AST is extremely sample efficient, but susceptible to mode collapse, where all discovered failures obey a single similar pattern. DiFS, while more computationally intensive, uncovers more likely and diverse failures.

%% file: Sections/methodology.tex
\section{Methodology}

We treat the driving stack as a black box that receives observations and outputs control actions, and at each timestep we inject a disturbance into the observations. The disturbance is additive noise on the perceived position of the nearest vehicle. A noise trajectory $\mathbf{x} = (a_1, \dots, a_T)$ contains the per-timestep disturbances $a_t$ over an episode of horizon $T$. Running the simulator with $\mathbf{x}$ produces a state trajectory, and we label the episode a failure if the truck collides with the nearby vehicle.

Let $p(\mathbf{x})$ be a disturbance model that assigns a probability density to each noise trajectory, reflecting how likely that trajectory is to occur for a given sensor system. Our goal is to sample the noise trajectories that cause a failure, weighted by their likelihood under $p(\mathbf{x})$:
\begin{equation}
p^{\star}(\mathbf{x} \mid \text{fail}) \propto
\mathbf{1}[\text{collision}]\, p(\mathbf{x})
\end{equation}
Direct Monte Carlo sampling from $p(\mathbf{x})$ rarely produces a collision because a reliable planner fails only for a small and unlikely set of trajectories. AST and DiFS are two strategies for concentrating samples on this failure region. Both take $p(\mathbf{x})$ as an input: AST rewards the log-probability $\log p(a_t)$ of each injected disturbance, and DiFS uses $p(\mathbf{x})$ as the prior it biases toward the failure distribution.

Real AV sensors have complex, skewed noise distributions that are expensive to characterize and specific to a given hardware configuration \cite{waymo}. Rather than model this noise empirically, we treat $p(\mathbf{x})$ as an upper bound on the perception system. This has two benefits. It removes the need to measure real sensor noise. It also bounds the discovered failures with respect to a safety specification rather than a particular sensor. A deployed sensor that meets the requirement injects disturbances no larger than those under $p(\mathbf{x})$, so its safety should be as good or better than simulated collisions found under this noise model.

We specify $p(\mathbf{x})$ as a zero-mean Gaussian whose standard deviation grows linearly with the longitudinal ($x$) and lateral ($y$) distance to the nearest actor vehicle, encoding the 
requirement that position accuracy may degrade with range. 

\pagebreak[4]
The defined distribution for this work is
\begin{equation}
\begin{aligned}
\mathbf{x} &\sim \mathcal{N}\!\left(\mathbf{0}, \Sigma(\mathbf{x})\right),
\quad
\Sigma(\mathbf{x}) =
\begin{bmatrix}
\sigma_x^2(x) & 0 \\
0 & \sigma_y^2(y)
\end{bmatrix}
\end{aligned}
\end{equation}

\begin{equation}
\begin{aligned}
\sigma_x(x) &= 0.02\,x + 1, \\
\sigma_y(y) &= 0.00625\,y + 0.2
\end{aligned}
\end{equation}

Both AST and DiFS solve the same underlying problem. AST uses reinforcement learning to adversarially search for failures in autonomous systems \cite{ritchie}. To use AST, we first formulate driving as a Markov decision process (MDP). Since the effects of sensor noise on driving behavior are stochastic, an MDP is a natural framework for this problem. Our goal is to teach an agent to add sensor disturbances that maximize the likelihood of a collision. During training, the agent concentrates its sampling in failure regions more than random Monte Carlo sampling. This bias makes failure discovery more sample-efficient by focusing exploration. 

\begin{figure}[H]
    \centering
    \includegraphics[width=\linewidth]{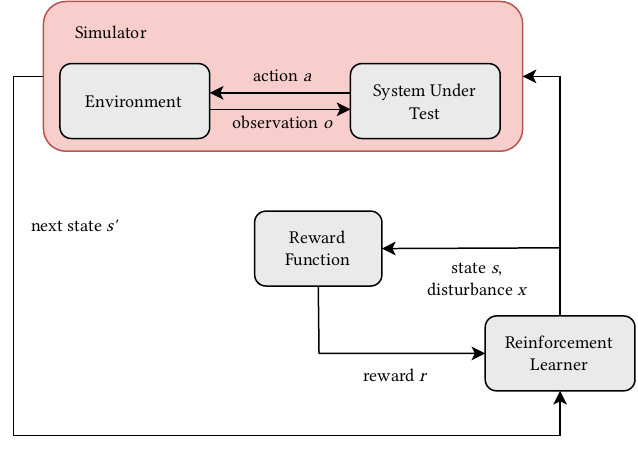}
    \caption{AST algorithm pipeline.}
    \label{fig:ast}
\end{figure} 

In an MDP, the reward function controls the agent's actions---here, it dictates what kind of sensor failures are found. To encourage likely collisions, we reward the log-probability of the injected noise $a$ at each timestep. Episodes not resulting in collision are penalized according to the minimum distance to the nearest vehicle observed over the entire episode, $d_{\min}$. Our reward function is given by
\begin{equation}
\begin{cases}
0,
& \text{if terminal failure}, \\[6pt]
-\alpha - c_{\text{dist}}\, d_{\min},
& \text{if terminal non-failure}, \\[6pt]
\log p(a), & \text{otherwise}
\end{cases}
\end{equation}
where $\alpha$ and $c_{\text{dist}}$ are tunable coefficients that penalize collisions and near-collisions. The reward function is customizable and can be augmented to include terms to direct the search (e.g. dissimilarity, Mahalanobis distance) \cite{ast_reward,ast_formulation}. We learn the agent's policy with the Soft Actor-Critic algorithm (SAC) to maximize reward while still allowing randomness. The choice of solver may affect the failure modes found \cite{ppo_sac}. 

DiFS trains a denoising diffusion model to recover increasingly dangerous noise profiles \cite{difs}. At each iteration, DiFS samples trajectories $\mathbf{x}$ from the current diffusion model $p(\mathbf{x}\mid r)$. The robustness $r$ of each noise trajectory is scored, defined here as the minimum distance to the nearest vehicle.  Samples above a robustness threshold $r_i$ are rejected, leaving only the least safe disturbances. The diffusion model is then retrained on these remaining samples. Because DiFS learns a distribution $p(\mathbf{x}\mid r)$, it excels at finding a diverse set of failures rather than isolated modes.

\begin{figure}[H]
    \centering
    \includegraphics[width=\linewidth]{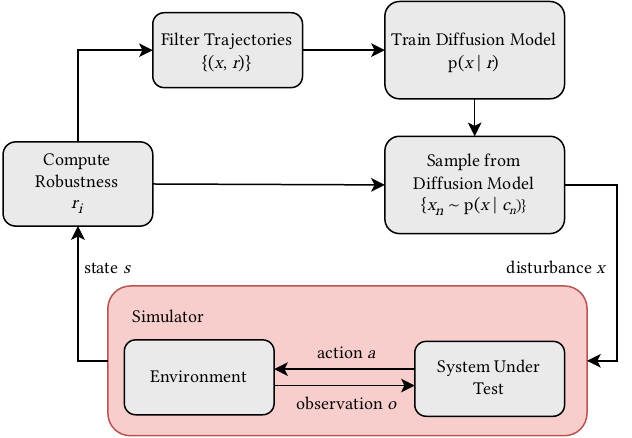}
    \caption{DiFS algorithm pipeline.}
    \label{fig:difs}
\end{figure}

The commercial planner samples and selects long-horizon paths according to a rules hierarchy, then optimizes them through quadratic programming for safety and comfort. The system also has safety overrides and distinct arrival, highway, and departure driving modes. This planner is integrated with Applied Intuition's Object Sim simulation software, which was selected for its cloud support and high-fidelity physics engine \cite{applied}. The commercial planner interfaces with Object Sim through ROS. We use ZeroMQ message passing to connect the AST/DiFS learning algorithm to the simulation environment and planner. This avoids modifying the planning software to enable black-box testing of the system.

\begin{figure*}[h!]
    \begin{center}
    \includegraphics[width=\linewidth]{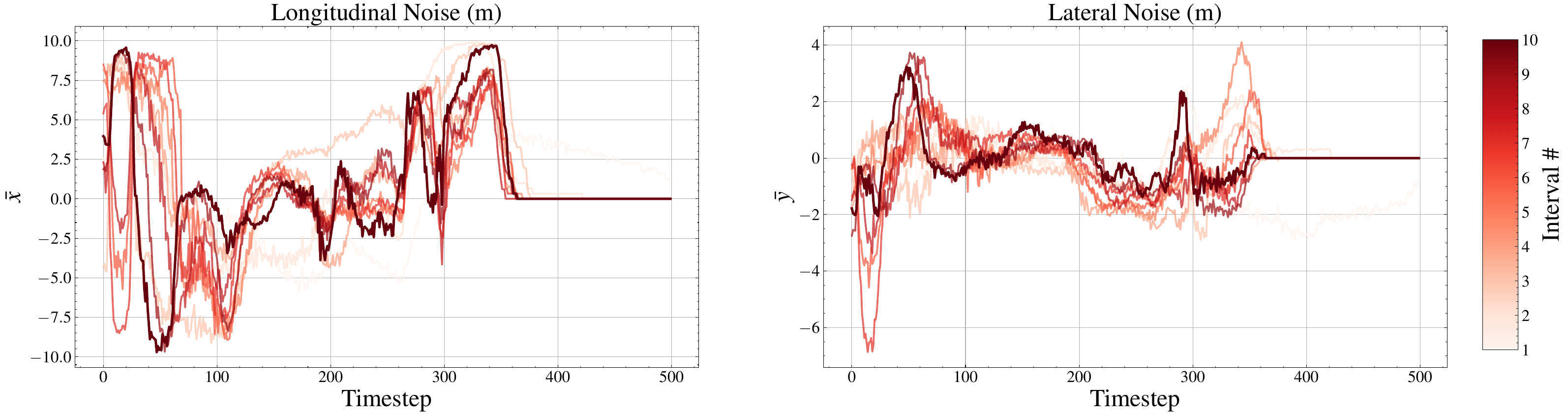}
    \caption{Disturbances found throughout AST training, averaged across 10 intervals. The adversarial\\ agent gradually learns noise profiles that cause the truck to collide with surrounding vehicles.}
    \label{fig:dist_over_time}
    \end{center}
\end{figure*}








Beyond discovering failures, we want to be able to characterize their structure. To analyze similarities among disturbance trajectories discovered by AST and DiFS, we perform principal component analysis (PCA) on the collected dataset. PCA transforms a large dataset of noise trajectories $D$ into a low dimensional representation $D^*$ while retaining maximal information. The singular value decomposition yields vectors in directions of maximum variance. These constitute the principal components, which are also the eigenvectors of the data's covariance matrix:
\begin{equation}
 D^* = DV_k,\quad\text{where}\quad D=U\Sigma V^T   
\end{equation}

We re-project $D$ using the first $k$ components (columns $V_k$). If a small number of components capture most of the variance, the trajectories are highly similar in structure \cite{uiuc}.
 
Principal components may vary throughout training as more likely and severe disturbance trajectories are discovered. We use K-means to cluster the set of principal components across all training episodes into distinct failure modes. Each principal component $\lambda$ is a vector indexed by timestep---we see empirically that the largest $\lambda$ entries coincide with moments just before simulated collision, where disturbances were most consequential. Principal component magnitudes therefore indicate which moments in a driving scenario are most sensitive to failures.

After clustering components in a latent space, we can invert the PCA transform (assuming mean $\mu$) and reconstruct trajectories in the original noise space using the following
\begin{equation}
\hat{D} = D^* V_k^T + \mu
\end{equation}

This yields a set of generalized noise  trajectories $\hat{D}$ that when re-applied to the same or similar driving scenarios reproduce failure modes observed over many episodes.

%% file: Sections/results.tex
\section{Results}

We compare the performance of our commercial vehicle intent planner to the baseline Intelligent Driver Model (IDM), a popular differential equation governing vehicle dynamics in traffic \cite{idm}. \Cref{tab:idm} shows that the commercial planner outperformed the IDM, yielding both a lower failure rate and lower average collision severity in Monte Carlo testing. Monte Carlo simulation represents one layer of a broader safety validation regime, and its inability to surface failures here reflects the planner's already high reliability rather than a gap in existing safety practices. Because traditional Monte Carlo cannot efficiently find failures in such a highly reliable system, the commercial planner presents a challenge, motivating targeted methods like AST and DiFS.

\input{tables/idm_stats}

Before deploying AST on the commericial planner, we performed a hyperparameter sweep to optimize AST performance. Due to testing environment and compute restrictions, the extent of our search was relatively limited. Each parameter was independently varied based upon heuristics from previous works \cite{ast_av}. Configurations were trained for 300 episodes. \Cref{tab:hyperparameters} shows the failure rate, average log-probability, and average severity. Severity is defined as the squared speed of the ego truck relative to the actor vehicle at collision time $\|v\|^2 = \|v_{\text{ego}} - v_{\text{actor}}\|^2$. Log-probability is the sum of the log noise probabilities at each timestep.

\input{tables/sweep}

Parameters $\alpha$ and $c_{\text{dist}}$ control the reward provided to the learning agent at the end of an episode---$\alpha$ penalizes non-collisions, and $c_{\text{dist}}$ penalizes near-collisions by the minimum distance from the ego truck to another vehicle. Batch size, buffer size, and learning rate are reinforcement learning parameters that influence training. A small batch size induces noisier, more frequent gradient updates. Buffer size is the maximum number of past episodes stored in memory. A SAC-specific value $\tau$ controls the target network update rate.

\input{tables/train_stats}
\input{tables/eval_stats}

We found that a small batch size ($\leq$16) universally improved failure rate, episode log-probability, and severity. A large buffer size and a small $\tau$ also significantly increased log-probability. A small $\alpha$ improved log-probability while shrinking failure rate, producing results more comparable to Monte Carlo testing. A large $c_\text{dist}$ increased failure rate and severity. Best-performing parameter values are marked bold in all three metrics of \Cref{tab:hyperparameters}. Parameter selection ultimately depends on which metric stakeholders prioritize.

DiFS is initialized with 500 Monte Carlo noise trajectories, trained until convergence, then 500 new trajectories are sampled and trained again. Finally, 500 new trajectories are evaluated. The DiFS quantile threshold is set to 0.3 to encourage fast convergence (the least robust 30\% noise trajectories are kept). A comparison of AST and DiFS performance in training is given in \Cref{tab:training_performance} and a comparison of Monte Carlo, AST, and DiFS in evaluation is given in \Cref{tab:evaluation_performance}. While longer training is preferable, our simulation environment limited feasible training time due to an upper bound on total simulation duration. Note that the cumulative log-probability of Monte Carlo simulations includes the entire simulation duration when no collision and early termination occurs, while the log-probability of AST or DiFS reflect the log probability only up to the time of collision. The compute cost to discover each failure is computed for a single AWS T4 GPU instance ($\$1.20$/hr).

As shown in \Cref{tab:evaluation_performance}, Monte Carlo simulation fails to find any failures in over 2000 episodes of training, while both AST and DiFS are able to find failures in just 300 episodes each. This demonstrates the effectiveness of these methods for safety validation of commercial driving stacks that meet high safety standards. Applying these methods helps quickly discover rare failures to continue improving overall system safety.

\begin{figure}[htb!]
    \centering
    \includegraphics[width=0.95\linewidth]{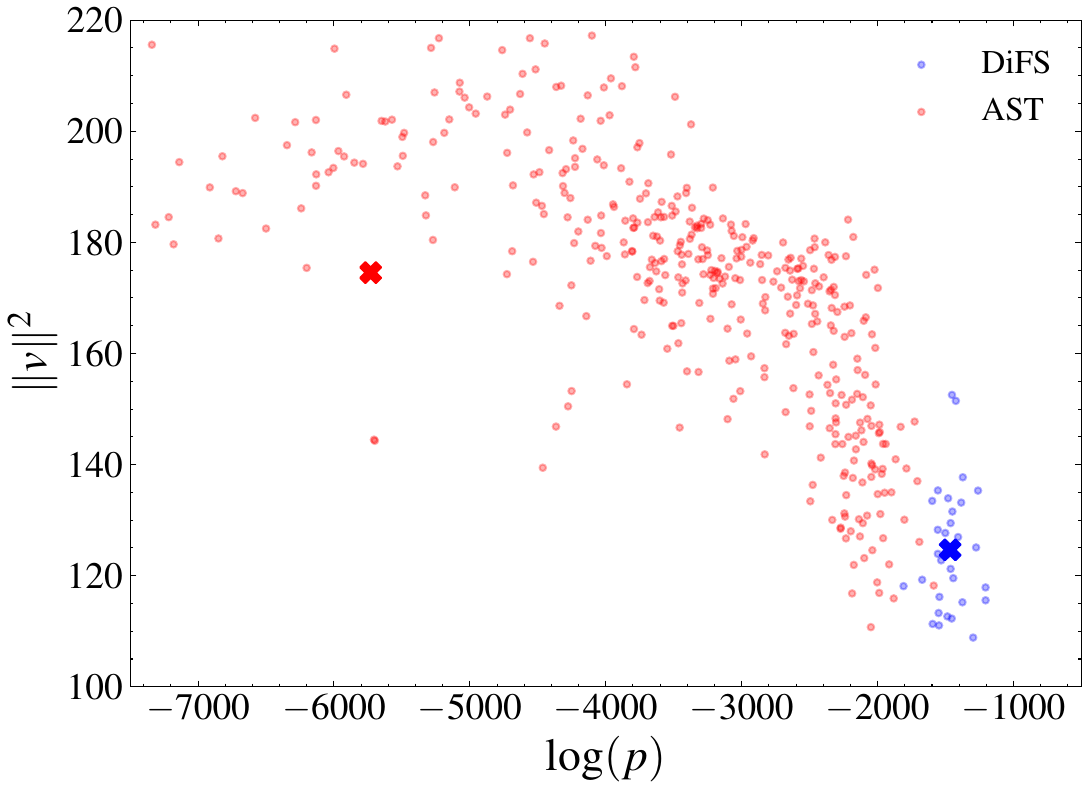}
    \caption{Likelihood vs. severity of AST and DiFS collisions.}
    \label{fig:sev_logprob}
\end{figure}

\Cref{fig:sev_logprob} shows the relationship between the log-probability and severity of AST and DiFS failures. As expected, simulated severe collisions are rarer than minor collisions. DiFS consistently finds the most likely failures, on par with Monte Carlo likelihoods.

To assess robustness to environmental variability, we evaluate our pre-trained AST policy on two variations of the base scenario used in training. Specifically, the truck's target cut-in distance was increased or decreased by 5 meters. The results of deploying the pre-trained AST policy on these new perturbed scenarios is shown in \Cref{tab:scenario_gen}. All 100 episodes resulted in collisions, showing the failure noise profile generalizes to other cut-in scenarios.

\input{tables/scenario_gen}
\footnotetext[1]{Refers strictly to cloud computing cost, computed for a single AWS T4 GPU instance (\$1.20/hr). It is not a valuation of collision events.}

Beyond merely finding collisions, AST and DiFS also find trajectories that violate other safety measures. We consider two measures here from the U.S. Department of Transportation: minimum time to collision (MinTTC) and the deceleration required to prevent a collision (DRAC) \cite{dept}. These failure measures are often used as early indicators of danger. They are derived from the vehicles' position and velocity at time $t$. We classify near-misses by MinTTC $<1.5$ s and unavoidable failures by DRAC $>7.0 \ \text{m}/\text{s}^2$, where recovery was physically impossible given the truck's peak braking capability. Collisions where DRAC $ \leq 7.0 \ \text{m}/\text{s}^2$ indicate planner failures.

Of 300 total DiFS trajectories, we observed nine collisions and ten non-collision near-misses. Of the nine collisions, four were unavoidable while five were planner failures. The average lead times between MinTTC and DRAC violations to collisions were 2.24 and 0.23 seconds, respectively. Further statistics are shown in \Cref{tab:ssms}.

\input{tables/ssms}
We demonstrate our PCA techniques on 300 failure episodes sampled from the pretrained AST policy. These methods can be similarly extended to DiFS failures, or to failure trajectories found by any validation algorithm.

\begin{figure}[h!]
    \centering
    \includegraphics[width=\linewidth]{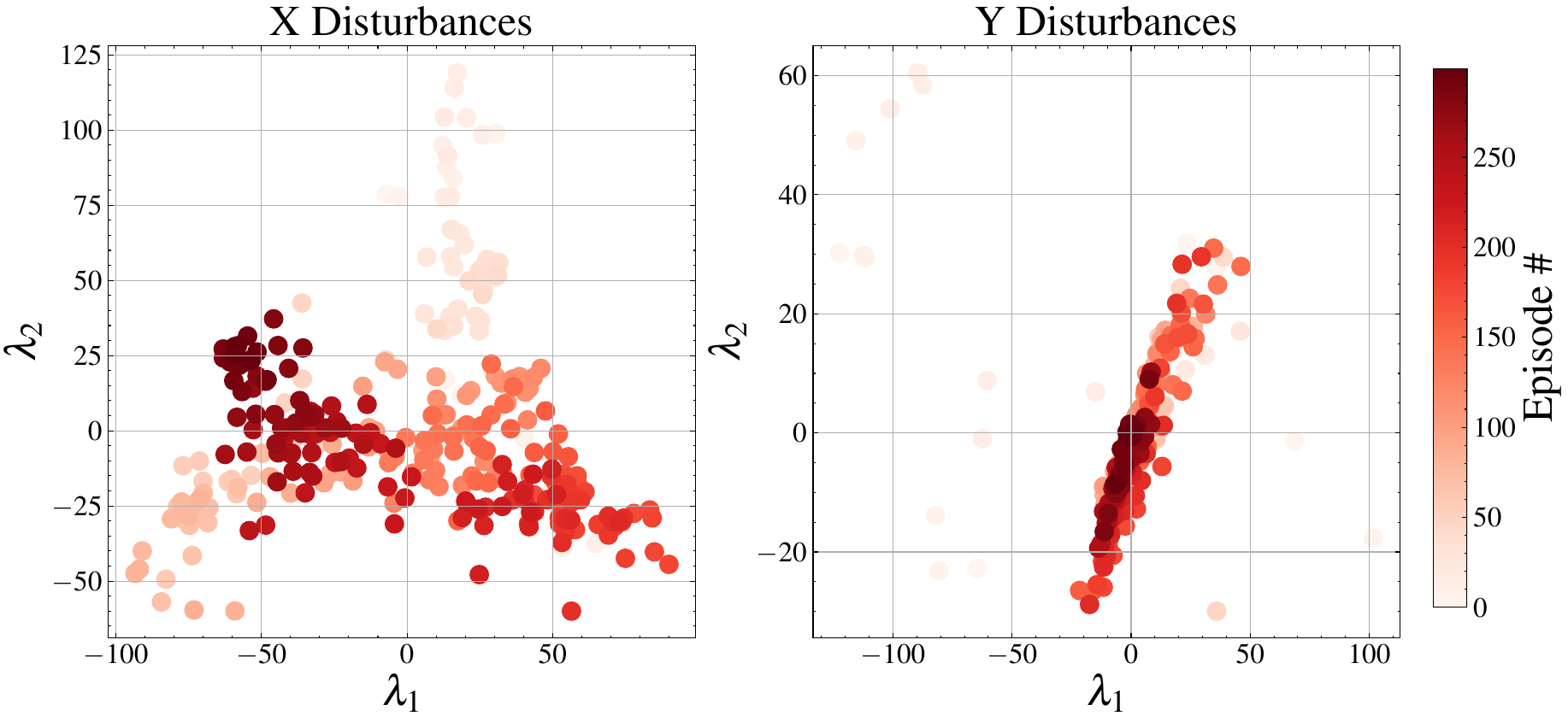}
    \caption{First and second principal components\\ of AST-generated failures.}
    \label{fig:principal_components}
\end{figure}

\Cref{fig:principal_components} shows the first two principal components throughout episodes. The lateral ($y$) eigenvalues are strongly linearly related, suggesting lateral disturbances follow a consistent, low-variance pattern across failures. The longitudinal ($x$) eigenvalues are generally negatively correlated, with higher variance and outlying groups. Clustering reveals three distinct failure modes. This correctly implies longitudinal noise has a larger impact on cut-in success than lateral sensor noise.

\begin{figure}[h!]
    \centering
    \includegraphics[width=\linewidth]{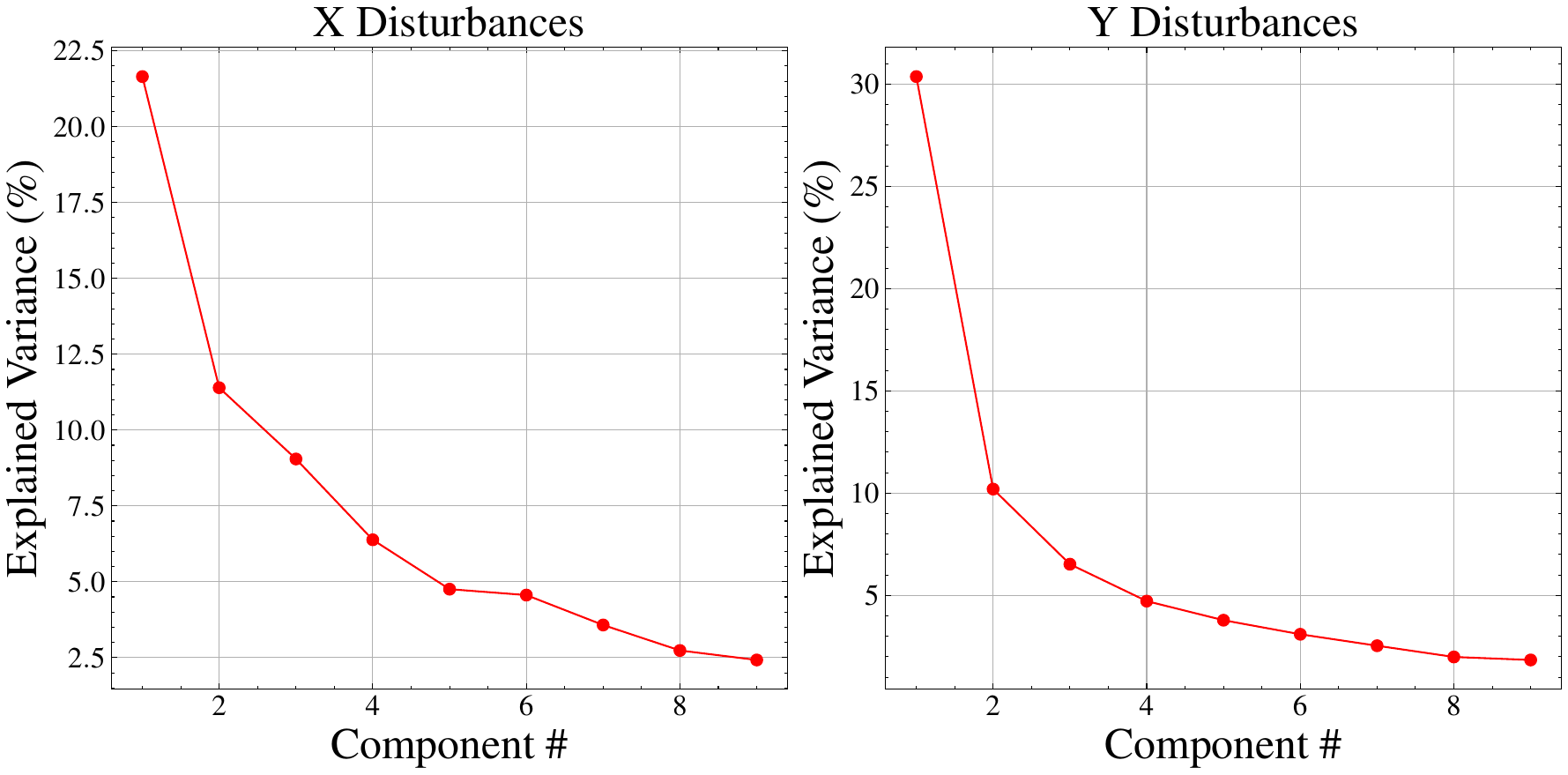}
    \caption{Variance explained by each principal component.}
    \label{fig:explained_variance}
\end{figure}

Explained variance is a measure of how effectively PCA uncovers a common low-dimensional trajectory. It corresponds directly to the eigenvalues of the data covariance matrix. \Cref{fig:explained_variance} indicates more principal components are needed to characterize longitudinal noise than lateral noise---the first three components only explain 40\% of the data variance. Future work should explore other noise model parameterizations, and whether similar low-dimensional structure emerges when analyzing heading or velocity error.

\begin{figure}[h!]
    \centering
    \includegraphics[width=\linewidth]{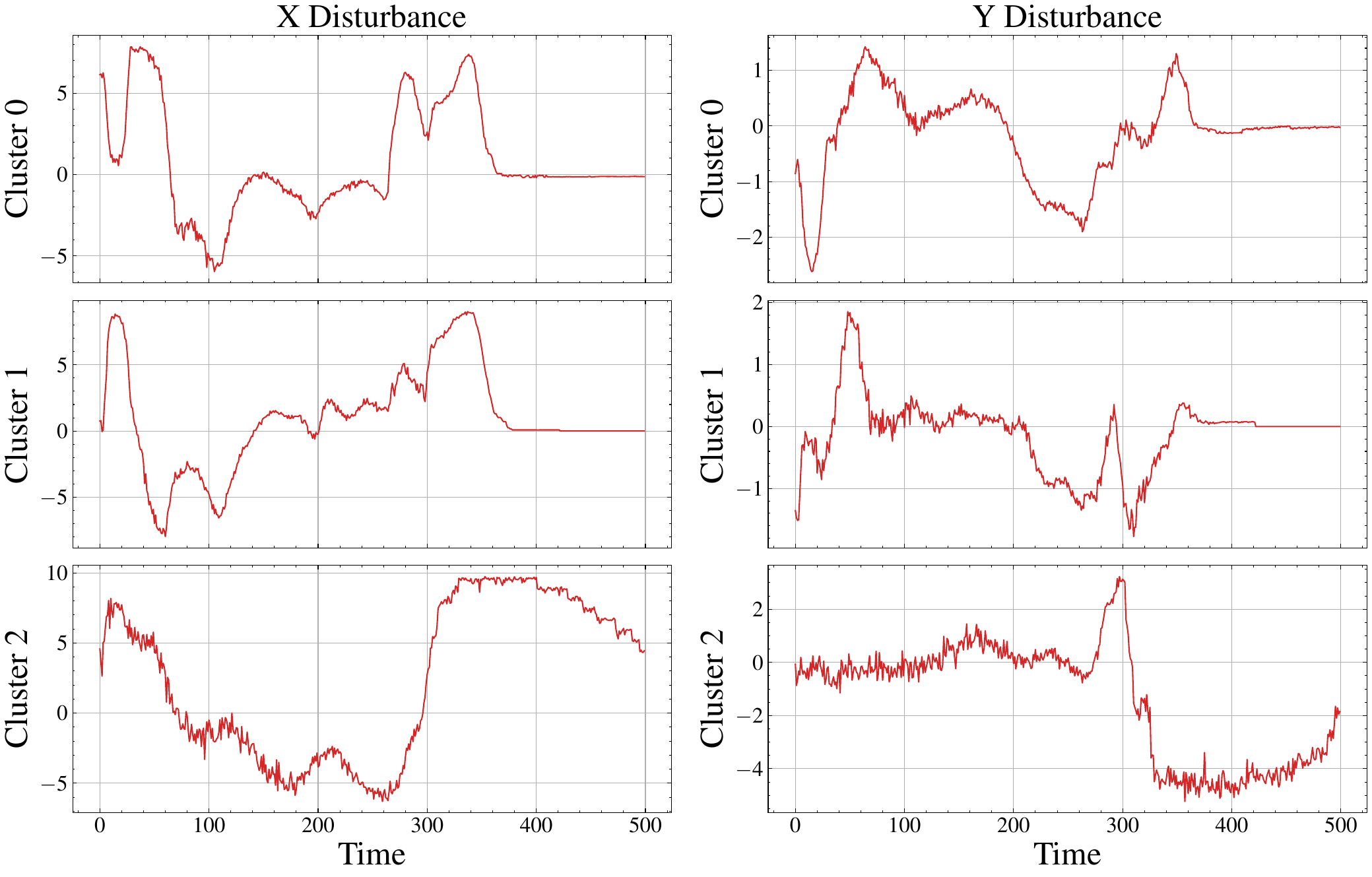}
    \caption{Generalized noise trajectories, or ``eigenfailures,"\\ recovered from inverting PCA.}
    \label{fig:recovered_trajectories}
\end{figure}

\Cref{fig:recovered_trajectories} presents the ``eigenfailures" recovered from inverting PCA on the three predominant clusters back into the original noise space. These eigenfailures reveal a weakness of AST and similar reinforcement learning methods: while a leftward bias in noise is necessary to induce a cut-in collision, AST attempts to maximize log-probability and reward by unnecessarily biasing the noise rightward before collision. Future studies should explore reward function adaptations or other interventions to mitigate this known weakness in reinforcement learning.

Importantly, we found that the PCA-derived noise trajectories reproduced failures in all scenarios tested, confirming they compactly represent failure modes of the planner. Of course, the generalization of these eigenfailures is limited by the sensor noise model accuracy. The sim-to-real gap remains unexplored in this work.

%% file: tables/idm_stats.tex
\begin{table}[h!]
\centering
\caption{Comparison of planners in Monte Carlo testing.}
\begin{tabular}{@{}lrr@{}}
\toprule
Planner & Failure Rate (\%) & Average Severity $\|v\|^2$\\
\midrule
Commercial &  0.0 &  0.000 \\
IDM  & 40.1 &  0.719 \\

\bottomrule
\end{tabular}
\label{tab:idm}
\end{table}

%% file: tables/sweep.tex
\begin{table}[H]
\centering
\caption{AST Hyperparameter Sweep}
\label{tab:hyperparameters}
\begin{tabular}{@{}llrrr@{}}
\toprule
Parameter & Value & Failure Rate (\%) & $\log(p)$ & $\|v\|^2$\\
\midrule
$\alpha$      & 500            & 45 & $-$2465.45 & 137.28 \\
              & 1000           & 72 & \textbf{$-$2124.35} & 126.37 \\
              & 5000$^\dagger$ & 55 & $-$3950.79 & 150.77 \\
              & 10000$^\ast$   & \textbf{96} & $-$5360.22 & \textbf{152.47} \\
\addlinespace
$c_{\text{dist}}$    & 100                   & 93 & $-$7340.49 & 145.40 \\
              & 1000$^\ast$$^\dagger$ & 96 & \textbf{$-$5360.22} & 152.47 \\
              & 5000                  & \textbf{98} & $-$7512.87 & \textbf{165.92} \\
\addlinespace
$\tau$        & 0.001$^\dagger$ & \textbf{99} & \textbf{$-$3218.86} & 136.15 \\
              & 0.005$^\ast$    & 96 & $-$5360.22 & 152.47 \\
              & 0.01            & 98 & $-$4002.00 & \textbf{155.10} \\
\addlinespace
Learning Rate & 0.001                   & 3 & \textbf{$-$1185.76} & 122.90 \\
              & 0.0001                  &  96 & $-$5694.37 & \textbf{188.23}  \\
              & 0.0003$^\ast$$^\dagger$ & \textbf{96} & $-$5360.22 & 152.47 \\
              & 0.00005                 & 3 & $-$1885.44 & 119.94 \\
\addlinespace
Batch size    & 16$^\dagger$ & \textbf{100} & \textbf{$-$4014.96} & \textbf{175.52}  \\
              & 64$^\ast$    & 96 & $-$5360.22 & 152.47 \\
              & 256          & 96 & $-$7250.31 & 165.64\\
\addlinespace
Buffer size  &  $10^4$$^\dagger$ & \textbf{98} & $-$3861.41 & 131.12 \\
             &  $10^5$$^\ast$    & 96 & $-$5360.22 & \textbf{152.47} \\
             &  $10^6$           & \textbf{98} & \textbf{$-$3182.21} & 148.69 \\
\bottomrule

\end{tabular}
\\[1ex]
\footnotesize
[$\ast$] Default value.
[$\dagger$] Selected configuration for main experiments.

\end{table}

%% file: tables/train_stats.tex
\begin{table*}[h!]
\centering
\caption{Training Performance}
\label{tab:training_performance}
\begin{tabular}{@{}llrrrrr@{}}
\toprule
Method & \# Episodes & Run Time & Failure Rate (\%) & GPU Instance Cost per Collision Episode (\$)\footnotemark[1] & Average Log-Probability & Average Severity $\|v\|^2$\\
\midrule
AST  & 1000 & 08:55:50 & 87.7 & 0.012 & $-$4506.13 & 152.63 \\
DiFS & 1000 & 12:57:11 & 2.4 & 0.648 & $-$1495.67 & 124.93 \\
\bottomrule
\end{tabular}
\end{table*}


%% file: tables/eval_stats.tex
\begin{table*}[h!]
\centering
\caption{Evaluation Performance}
\label{tab:evaluation_performance}
\begin{tabular}{@{}llrrrrr@{}}
\toprule
Method & \# Episodes & Run Time & Failure Rate (\%) & GPU Instance Cost per Collision Episode (\$)\footnotemark[1] & Average Log-Probability & Average Severity $\|v\|^2$\\
\midrule
MC   & 2000 & 26:14:29 & 0.0 & N/A & N/A & N/A \\
AST  & 300 & 05:32:08 & 94.6 & 0.023 & $-$2870.26 & 164.75\\
DiFS & 300 & 03:53:10 & 3.1 & 0.518 & $-$1490.71 & 115.21\\
\bottomrule
\end{tabular}
\end{table*}

%% file: tables/scenario_gen.tex
\begin{table}[H]
\centering
\caption{AST Scenario Generalization Performance}
\label{tab:scenario_gen}
\begin{tabular}{@{}llrr@{}}
\toprule
Scenario & \# Episodes & Failure Rate (\%) & Average $\|v\|^2$\\
\midrule
Cut-in      & 300 & 94.6  & 164.75 \\
Cut-in ($+$5) & 100 & 100.0 & 190.50 \\
Cut-in ($-$5) & 100 & 100.0 & 174.45 \\

\bottomrule
\end{tabular}
\end{table}

%% file: tables/ssms.tex
\begin{table}[H]
\centering
\caption{Safety Measures of 300 DiFS Episodes}
\label{tab:ssms}
\begin{tabular}{@{}lrrr@{}}
\toprule
Metric & Count (of 300) & Median & Worst Case\\
\midrule
Collision          & 9   & N/A  & N/A \\
MinTTC $< 1.5$ s   & 19  & 3.50 s & 0.015 s\\
DRAC $> 7$ m/s²    & 4   & 0.40 m/s² & 43.633 m/s²\\
\bottomrule
\end{tabular}
\end{table}

%% file: Sections/conclusion.tex
\section{Conclusion}

We have presented two promising techniques for rare-event failure discovery that scale to commercial AV planners, which exhibit dramatically lower failure rates than standard driver models. By performing PCA on AST and DiFS results, we uncover the eigenfailures, the underlying canonical failure modes of the system. These reproduce collisions when applied to identical and similar driving scenarios.

Immediate future work includes training both AST and DiFS on a larger variety of scenarios to prevent overfitting and characterize more failure conditions. Techniques to address and prevent mode collapse would also be beneficial. One method is to dynamically rescale the noise model throughout AST and DiFS training, allowing greater control over the convergence rate. Ultimately, eigenfailure analysis offers a path toward not just discovering autonomous system failures, but diagnosing causes.

%% file: Sections/acknowledgement.tex
\section*{Acknowledgments}

This work was supported by Torc Robotics through the Stanford Center for AI Safety, the Stanford Graduate Fellowship program, and the Stanford Institute for Human-Centered AI.

%% file: references.bib
@String { iccv        = {International Conference on Computer Vision (ICCV)} }

@String { itsc        = {IEEE International Conference on Intelligent Transportation Systems (ITSC)} }

@String { iv          = {IEEE Intelligent Vehicles Symposium (IV)} }

@String { jair        = {Journal of Artificial Intelligence Research} }

@String { neurips     = {Advances in Neural Information Processing Systems (NeurIPS)} }

@String { rss         = {Robotics: Science and Systems} }

@inproceedings{ast_av,
  author={Koren, Mark and Alsaif, Saud and Lee, Ritchie and Kochenderfer, Mykel J.},
  title={Adaptive Stress Testing for Autonomous Vehicles}, 
  booktitle=iv, 
  year={2018},
  pages={1--7},
  url={https://ieeexplore.ieee.org/abstract/document/8500400}
}

@inproceedings{difs,
  author={Delecki, Harrison and Schlichting, Marc R. and Arief, Mansur and Corso, Anthony and Vazquez-Chanlatte, Marcell and Kochenderfer, Mykel J.},
  booktitle={IEEE Engineering Reliable Autonomous Systems (ERAS)}, 
  title={Diffusion-Based Failure Sampling for Evaluating Safety-Critical Autonomous Systems}, 
  year={2025},
  pages={1--8}
}

@inproceedings{ast_reward,
  author={Corso, Anthony and Du, Peter and Driggs-Campbell, Katherine and Kochenderfer, Mykel J.},
  booktitle=itsc, 
  title={Adaptive Stress Testing with Reward Augmentation for Autonomous Vehicle Validation}, 
  year={2019},
  pages={163--168},
  doi={10.1109/ITSC.2019.8917242}
}

@inproceedings{ast_formulation,
    author={Koren, Mark and Corso, Anthony and Kochenderfer, Mykel J.},
    booktitle={Workshop on Safe Autonomy, Robotics: Science and Systems (RSS)},
    title={The Adaptive Stress Testing Formulation},
    year={2019}
}

@article{pca_bn,
    author={Adedigba, Sunday A. and Khan, Faisal and Yang, Ming},
    journal={Industrial \& Engineering Chemistry Research},
    title={Dynamic Failure Analysis of Process Systems Using Principal Component Analysis and Bayesian Network},
    year={2017},
    volume={56},
    number={18},
    pages={2094–-2106},
    doi={10.1021/acs.iecr.6b03356}
}

@inproceedings{rocas,
  author    = {Feng, Shiwei and Ye, Yapeng and Shi, Qingkai and Cheng, Zhiyuan and Xu, Xiangzhe and Cheng, Siyuan and Choi, Hongjun and Zhang, Xiangyu},
  title     = {{ROCAS}: Root Cause Analysis of Autonomous Driving Accidents via Cyber-Physical Co-mutation},
  booktitle   = {International Conference on Automated Software Engineering},
  pages     = {1620--1632},
  year      = {2024},
  doi       = {10.1145/3691620.3695530},
}

@article{kurse,
	title = {Experimental determination of factors causing crashes involving automated vehicles},
	doi = {doi.org/10.1016/j.multra.2024.100186},
	journal = {Multimodal Transportation},
	author = {Kurse, Teshome Kumsa and Gebresenbet, Girma and Daba, Geleta Fikadu and Tefera, Negasa Tesfaye},
	year = {2025},
	number = {100186},
    pages = {1--18}
}

@book{uiuc,
  author    = {Ian T. Jolliffe},
  title     = {Principal Component Analysis},
  year      = {2002},
  edition   = {2},
  publisher = {Springer},
  address   = {New York},
}

@Article{ritchie,
  author  = {Ritchie Lee and Ole J. Mengshoel and Anshu Saksena and Ryan W. Gardner and Daniel Genin and Joshua Silbermann and Michael Owen and Mykel J. Kochenderfer},
  journal = jair,
  title   = {Adaptive stress testing: finding likely failure events with reinforcement learning},
  year    = {2020},
  pages   = {1165--1201},
  volume  = {69},
  doi     = {10.1613/jair.1.12190},
}

@article{ppo_sac,
    author = {Mock, J. and Muknahallipatna, S.},
    title = {Sim-to-Real: A Performance Comparison of {PPO}, {TD3}, and {SAC} Reinforcement Learning Algorithms for Quadruped Walking Gait Generation},
    journal = {Journal of Intelligent Learning Systems and Applications},
    year = {2024},
    volume={16},
    number={2},
    pages = {23--43},
    doi = {10.4236/jilsa.2023.151003},
}

@techreport{moss,
title = {Autonomous Vehicle Risk Assessment},
author = {Moss, Robert J. and Gupta, Shubh and Dyro, Robert and Leung, Karen},
institution = {Stanford University, Department of Aeronautics and Astronautics},
year = {2021},
month = {June},
url = {https://robert-moss.com/pdf/av_risk_assessment.pdf}
}

@article{idm,
  title = {Congested traffic states in empirical observations and microscopic simulations},
  author = {Treiber, Martin and Hennecke, Ansgar and Helbing, Dirk},
  journal = {Physical Review E},
  volume = {62},
  pages = {1805--1824},
  year = {2000},
  month = {Aug},
  publisher = {American Physical Society},
  doi = {10.1103/PhysRevE.62.1805},
  url = {https://link.aps.org/doi/10.1103/PhysRevE.62.1805}
}

@misc{applied,
    author={{Applied Intuition Inc.}},
    title={{Object Sim}},
    howpublished={Mountain View, CA, USA. \url{www.appliedintuition.com/products/object-sim}},
    year={2026},
}

@article{waymo,
  title     = {Large Scale Interactive Motion Forecasting for Autonomous Driving: The {Waymo} Open Motion Dataset},
  author    = {Ettinger, Scott and others},
  journal = {International Conference on Computer Vision (ICCV)},
  year      = {2021},
  month     = {Oct},
  pages     = {9710--9719},
}

@article{koopman,
  title = {Challenges in Autonomous Vehicle Testing and Validation},
  author = {Koopman, P. and Wagner, M.},
  journal = {SAE International Journal of Transportation Safety},
  year={2016},
  month={Apr},
  doi={10.4271/2016-01-0128},
  }

@article{neurips,
 author = {Sinha, Aman and O\textquotesingle Kelly, Matthew and Tedrake, Russ and Duchi, John C},
 journal = {Conference on Neural Information Processing Systems},
 pages = {6402--6416},
 title = {Neural Bridge Sampling for Evaluating Safety-Critical Autonomous Systems},
 year = {2020}
}

@article{dept,
    title = {{Surrogate Safety Measures From Traffic Simulation Models Final Report Publication No: FHWA-RD-03-050}},
    year = {{\textit{U.S. Department of Transportation Federal Highway Administration}}
    }
    }
